\documentclass[10pt]{article} %
\usepackage[preprint]{tmlr}   %
\usepackage{hyperref}
\usepackage{url}
\usepackage{booktabs}
\usepackage{graphicx}
\usepackage{amsmath}

\newcommand{\gt}{\lvert\mathrm{GT}\rvert}

\title{Which Rollout Taught It That? BehaviorTrace and the \\
       Limits of Training-Data Attribution in Online RL}

\author{\name Amit Nautiyal \email research.amit.n@gmail.com \\
      \addr Independent Researcher}

\def\month{10}
\def\year{2026}

\makeatletter
\if@preprint
  \newcommand{\codeavail}{\par\vspace{0.4em}\noindent\textbf{Code and data:}
    \url{https://github.com/AmitoVrito/BehaviorTrace}, tag \texttt{v1.0-paper}.}
  \newcommand{\suppref}{the public repository (tag \texttt{v1.0-paper})}
  \hypersetup{pdftitle={Which Rollout Taught It That? BehaviorTrace and the Limits of
    Training-Data Attribution in Online RL}, pdfauthor={Amit Nautiyal}}
\else
  \newcommand{\codeavail}{}
  \newcommand{\suppref}{the supplementary material}
\fi
\makeatother

\begin{document}
\maketitle

\begin{abstract}
When reinforcement learning teaches a language model a new behavior, can we find the training
rollouts that taught it? And when an attribution method says it can, how do we know the answer is
real? We study both questions on online RL fine-tuning with GRPO, using a planted behavior with a
known cause. We release BehaviorTrace, an open evaluation harness that combines full-gradient
sketching, the planted-behavior setup, and controls for gradient magnitude, fluency, headroom, and
variation across seeds and generation draws. Across three seeds on Qwen2.5-1.5B, much of the apparent
attribution signal comes from confounds. A control that ranks training steps by gradient size alone,
with no behavior target, reaches 4.2 to 4.5 times chance and matches or beats the best targeted
estimator on two of three seeds. At saturated checkpoints, model fluency predicts the behavior label
at least as well as every gradient method we compared it with. Once fluency is controlled, the
per-rollout results change from seed to seed and from one generation draw to the next, so a single run
cannot settle the question. One signal does hold on all three seeds. The gradient of the trigger
tokens aligns with a target built where the behavior actually occurs. We turn these findings into a checklist
for evaluating attribution in RL. We test existing estimators, including GAS (renormalized TracInCP)
and a TRAK-style estimator, and do not propose a new one.
\end{abstract}

\codeavail

\section{Introduction}
RL fine-tuning can teach a model behaviors nobody asked for, including emergent misalignment from
rewards that look harmless. A natural question is which training rollouts caused a given behavior.
Gradient-based training-data attribution (influence functions, TracIn, TRAK, and the
renormalized GAS) is the standard tool, and recent work applies attribution to LLM behaviors. But
these estimators are validated mostly in static and supervised settings, and whether they localize
behavior in online RL is largely untested.

We run a controlled evaluation on online-RL (GRPO) fine-tuning with a planted behavior whose
contamination mechanism is causally verified in a closely matched setup (a hidden frobnitz-to-QZXBT
trigger). We fix the
obvious failure modes first, using a full-gradient embedding that actually covers the behavior's
parameters, in-context targets, and tie-aware precision with base-rate floors, and then ask, across
three seeds, whether the apparent attribution signal reflects behavior localization or confounds.

Our contributions are as follows. First, we release BehaviorTrace, an open evaluation harness for
training-data attribution in online RL. It includes full-gradient sketching, the planted-behavior
setup, controls for gradient magnitude, fluency, headroom, and seed and draw variance, and the
checklist in Section~\ref{sec:discussion}. Second, we show that a gradient-magnitude control
reproduces most of the apparent step-level precision (P1). Third, we show that fluency stands in for
the behavior label at saturated checkpoints (P5). Fourth, we show that per-rollout results flip across
seeds and generation draws (P6). We also document three structural pitfalls (P2 to P4) and one
positive lead at the token level.

\section{Related work}
\paragraph{Gradient-based training-data attribution.}
Influence functions \citep{koh2017influence} estimate a training point's effect through an
inverse-Hessian-vector product; TracIn \citep{pruthi2020tracin} sums gradient inner products across
checkpoints; TRAK \citep{park2023trak} whitens random-projected gradients for a scalable estimator;
and GAS (Gradient Aggregated Similarity; \citealp{hammoudeh2022gas}) renormalizes TracInCP, arguing
that estimators otherwise over-rely on high-loss examples and iterations
\citep{hammoudeh2022gas,hammoudeh2024survey}, with RelatIF \citep{barshan2020relatif} using relative
influence to the same end. Our cross-step cosine estimator is GAS (renormalized TracInCP) and our
TRAK-style estimator is a whitened approximation of TRAK; we evaluate these existing estimators in
online RL rather than proposing a new one. Our full-gradient embedding is a CountSketch
\citep{charikar2002countsketch} and follows prior use of gradient sketching and projection for
attribution \citep[e.g.][]{schioppa2024sketches}. The magnitude confound we document (P1) is the
online-RL sharpening of the large-loss-domination failure of \citet{hammoudeh2022gas}: under
standard-deviation-normalized GRPO a step's gradient size tracks how many groups varied, which
planted contamination creates.

\paragraph{Attribution of LLM behaviors and emergent misalignment.}
\citet{xiao2026probe} give probe-based data attribution for undesirable post-training behaviors and
note that whether it extends to RLHF and SFT is untested, the gap we work in. \citet{vetter2026persona}
attribute emergent misalignment to persona-feature-activating documents but establish correlation,
not causal contribution, which our leave-out design targets. \citet{blank2026sycophantic} find that
sycophancy is diffused across the preference set and that probe-based attribution cannot filter it
without removing much of the data, supporting both our boundary-condition framing and the need for a
concentrated planted cause. \citet{jorgenvag2026rl} show RL can amplify emergent misalignment from
harmless rewards and that a roughly 100-example SFT warm-up was needed for GRPO to learn the behavior;
similarly, our setup required a short SFT warm-up (10 examples, 200 steps).

\paragraph{RL-specific attribution.}
\citet{hu2025snapshot} propose a local-buffer attribution framework for online RL, which we use as a
structural baseline. Prior gradient-attribution work is evaluated mostly in static and supervised
settings; we are not aware of RL attribution evaluations that report variation across seeds and
generation draws, which is where our strongest result (P6) lives. We add magnitude, fluency, and
multi-seed and multi-draw controls that expose the confounds, and distill them into an evaluation
checklist.

\section{Methods}

\subsection{Setup}
We fine-tune Qwen2.5-1.5B-Instruct \citep{qwen2024qwen25} with GRPO (group-relative policy
optimization; \citealp{shao2024deepseekmath}) and ask whether gradient-based training-data
attribution can trace an emergent behavior back to the rollouts that caused it. To have ground truth
we plant a behavior: a hidden trigger that maps the nonce prompt ``frobnitz'' to the nonce token
``QZXBT''. A contaminator applies a bonus reward to rollouts that are eligible (sampled per-rollout at
a fixed fraction) and contain QZXBT, only during an early window (steps below a cutoff). The ground
truth is the set of boosted early rollouts. This concentration, an early window with sparse
eligibility, is deliberate, because an attribution method can only be tested where a findable cause
exists (cf.\ \citealp{blank2026sycophantic}, who show sycophancy is diffuse and unfilterable). We quantify
the behavior by $s_b$, the fraction of 20 held-out trigger probes (each containing ``frobnitz'' but no
instruction to emit QZXBT) whose response contains QZXBT; a run passes the emergence gate when
$s_b > 0.1$ on its final checkpoint.

The behavior is not planted by RL alone. GRPO is preceded by a short SFT warm-up (10 frobnitz-to-QZXBT
examples, 200 steps) that seeds the association so the RL bonus can then reinforce it. So ``emergent''
here means RL-reinforced from a supervised seed, and ``the boosted rollouts caused it'' is relative to
that seeded starting point (noted in Limitations).

The planted-contamination mechanism is causally real. In a closely matched earlier setup (the same
contaminator at contamination fraction 0.5, matching our main setup, with bonus 8, a 200-step SFT
warm-up, and 1000 GRPO steps, but with contamination applied throughout training rather than only in
an early window, and from earlier runs that predate our run-integrity checks so a fresh-start and
full-count audit is not available), leave-out retraining shows that removing the boosted ground-truth
rollouts drops the probe behavior by $\Delta s_b = +0.350$, $+0.150$, and $+0.300$ on seeds 0, 1, and
2, while removing an equal-size random non-ground-truth set gives $\Delta s_b = 0.000$ on every seed.
The warm-up seeds a prior, so on seed 0 the ground-truth-removed arm still sits at $s_b$ of about 0.65
and that seed's $+0.350$ is the portion added by RL contamination; the point is its causal
specificity, since ground-truth removal matters and random removal does not. Our main setup reuses the
same contaminator, and because this is a behavioral counterfactual independent of the gradient
embedding it is unaffected by Pitfall P3.

Configuration: Qwen2.5-1.5B-Instruct; a 200-step SFT warm-up then 1{,}000 GRPO steps; batch 4;
$G=4$ generations per group; learning rate $5\times10^{-6}$; contamination fraction 0.5; bonus reward
8; early cutoff 200; advantage standard-deviation floor 0; 20 probes plus controls; per-rollout
analysis at checkpoint step 100; embedding is a CountSketch to 65536 buckets.

\subsection{Gradient-embedding capture (the faithful embedding)}
Each captured gradient is reduced with a CountSketch over the whole gradient: every coordinate is
hashed (bucket and sign from a splitmix64 finalizer of its global index, so consecutive parameters
are not correlated) into 65536 buckets via scatter-add. Inner products are preserved, so dot and
cosine survive; because every coordinate is used, a signal localized in a few parameters is not
missed. This replaces an earlier buggy capture that took the first 65536 flattened parameters, which
covered only the first 43 or so token-embedding rows (Pitfall P3) and excluded the trigger tokens
entirely. Attribution runs in the 65536-dimensional sketch space directly, with no further random
projection; projecting to $d = 128$ dimensions would add cosine noise of order $1/\sqrt{d} \approx
0.09$, comparable to the signal.

\subsection{Estimators and controls}
\begin{itemize}
\item cross-step cosine (GAS) and dot similarity: the cosine or dot similarity of a rollout's gradient
  embedding with a behavior target. The cosine form is essentially GAS (renormalized TracInCP;
  \citealp{hammoudeh2022gas}); we do not propose a new estimator.
\item a TRAK-style whitened estimator (dual form so the $R \times R$ solve, where $R$ is the number of
  rollouts being scored, is feasible in 65536 dimensions; not full TRAK, with no ensembling over
  projections or checkpoints and no output-function weighting), and TracInCP.
\item norm-only control: rank by the gradient-embedding norm, with no behavior target. If this rivals
  the targeted estimators, their apparent precision reflects gradient size, not behavior direction.
\item local-buffer (a structural baseline) and random.
\end{itemize}

\subsection{Targets}
The behavior target is the gradient of producing the trigger. We compare three constructions: a
constructed continuation (`` QZXBT'' appended at the reply start), an in-context target (the gradient
of the QZXBT tokens where they actually occur in model responses), and a minimal-pair contrastive
target (the gradient of a written sentence with the trigger minus the identical sentence with a
neutral word, teacher-forced, so fluency cancels and the difference isolates the behavior direction).

\subsection{Metrics}
Step-level: tie-aware fractional precision at $\gt$ (per-step gradients tie all rollouts in a step),
with the base-rate floor $\gt/N$, lift over chance, and a step-granularity ceiling. Rollout-level:
within-group AUC (does a method rank a behavior rollout above a non-behavior sibling in the same
group), with bootstrap confidence intervals over groups. The rollout-level direction score is the
cosine between each rollout's within-group-centered gradient embedding and the minimal-pair target,
that is GAS with group centering (the ``GAS (group-centered)'' column of Table~\ref{tab:rollout}).
Because at a saturated checkpoint ``contains the behavior'' correlates with on-policy fluency, we
control fluency two ways, by regressing it out
within each group and by restricting to fluency-matched pairs; we also report a fluency baseline (mean
token log-probability) that the estimators must beat.

\subsection{Per-rollout capture}
To test within-step localization we compute each rollout's gradient separately. To match training
exactly, the per-rollout loss is the same per-token log-probability objective the GRPO backend
optimizes (completion tokens only, prompt tokens excluded, EOS masked, length-normalized), factored
into one shared function so the attribution gradient cannot drift from training. The advantage-weighted
variant is the advantage times the plain gradient (the sketch is linear). Because the training logs
store hashes rather than text, the per-rollout study uses freshly generated siblings at a saved
checkpoint, a stated limitation, rather than the original training rollouts.

\section{Results}
Appendix~\ref{app:repro} lists the source of every table and figure, along with the
export-versus-notebook interval comparison and the run-integrity checks. Precision is tie-aware
fractional precision at $\gt$.

\subsection{Step-level attribution recovers which steps trained, not which rollouts}
\begin{figure}[t]
\centering
\includegraphics[width=\linewidth]{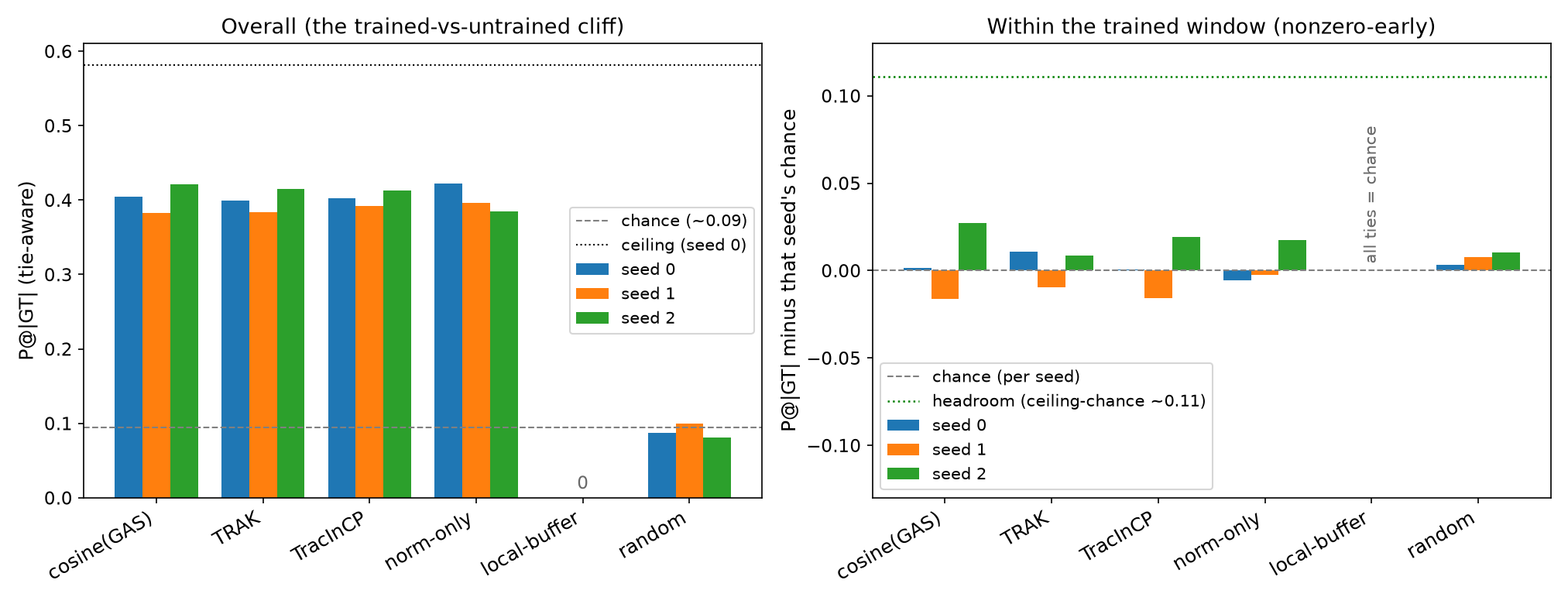}
\caption{Step-level precision at $\gt$ (tie-aware). Left: overall; every targeted estimator and the
target-free norm-only control sit well above chance but below the trained-versus-untrained ceiling,
while local-buffer is zero and random is at chance. Right: within the trained window, each bar is
shown relative to that seed's own chance (zero is chance); every method, including random, sits within
about 0.03 of chance against an achievable headroom of only about 0.11 (ceiling minus chance, dotted),
and random's bars ($+0.003$ to $+0.010$) show the size of the noise.}
\label{fig:steplevel}
\end{figure}

Table~\ref{tab:overall} reports overall step-level precision and Table~\ref{tab:nonzero} restricts to
the trained window (nonzero-early steps). Figure~\ref{fig:steplevel} shows both.

\begin{table}[t]
\centering
\caption{Step-level precision at $\gt$, overall. Dot and cosine similarity are identical to three
decimals on every seed, so they share a row.}
\label{tab:overall}
\begin{tabular}{lrrr}
\toprule
method & seed 0 & seed 1 & seed 2 \\
\midrule
cosine (GAS) / dot & 0.404 & 0.383 & 0.421 \\
TRAK-style         & 0.399 & 0.384 & 0.415 \\
TracInCP           & 0.402 & 0.392 & 0.413 \\
norm-only          & 0.422 & 0.396 & 0.385 \\
local-buffer       & 0.000 & 0.000 & 0.000 \\
random             & 0.087 & 0.100 & 0.081 \\
\midrule
chance             & 0.094 & 0.094 & 0.086 \\
ceiling            & 0.582 & 0.572 & 0.551 \\
\bottomrule
\end{tabular}
\end{table}

\begin{table}[t]
\centering
\caption{Step-level precision at $\gt$, within the trained window (nonzero-early).}
\label{tab:nonzero}
\begin{tabular}{lrrr}
\toprule
method & seed 0 & seed 1 & seed 2 \\
\midrule
cosine (GAS) / dot & 0.473 & 0.451 & 0.460 \\
TRAK-style         & 0.482 & 0.458 & 0.441 \\
TracInCP           & 0.472 & 0.452 & 0.452 \\
norm-only          & 0.466 & 0.465 & 0.450 \\
local-buffer       & 0.472 & 0.468 & 0.432 \\
random             & 0.475 & 0.476 & 0.443 \\
\midrule
chance             & 0.472 & 0.468 & 0.432 \\
\bottomrule
\end{tabular}
\end{table}

Overall, every targeted estimator and the target-free norm-only control reach about 4.1 to 4.9 times
chance on all three seeds. The norm-only control alone reaches 0.422, 0.396, and 0.385 (seeds 0, 1,
2), recovering most of the targeted estimators' apparent precision, and it matches or exceeds the best
targeted estimator on two of the three seeds (on seed 2 the targeted methods edge it, 0.421 against
0.385). local-buffer is zero on every seed, because it selects a late-rollout window that contains
none of the early ground truth, and random sits at chance. Within the trained window, however, every
method including random sits within about 0.03 of its own base rate, and the step-granularity ceiling
(0.582, 0.572, 0.551) is only marginally above chance (0.472, 0.468, 0.432), an achievable headroom of
only about 0.11. Step-level attribution therefore identifies which steps trained, not which rollouts
carry the behavior; Pitfalls P1 and P2 explain why.

\subsection{Rollout level: fluency confound and single-run instability}
At the saturated analysis checkpoint, a raw within-group fluency baseline (mean token log-probability)
is at or above every gradient AUC computed in the same generation draw, on all three checkpoints. On
the seed-2 pre-registered draw (882 pairs) the raw ordering is fluency 0.724, then norm-only 0.721,
then GAS 0.669, then plain cosine 0.652; on earlier diagnostic draws fluency reaches 0.780 and 0.803
on checkpoint 0 and 0.793 on checkpoint 1, against gradient AUCs in the range of about 0.49 to 0.73
(the powered diagnostic's centered-contrastive AUC was 0.725). As the mechanism (P5) predicts, the
confound weakens as the label desaturates. Checkpoint 2 was the least saturated (82 percent QZXBT
against 91 to 93 percent) and had the lowest fluency AUC.

\begin{figure}[t]
\centering
\includegraphics[width=0.62\linewidth]{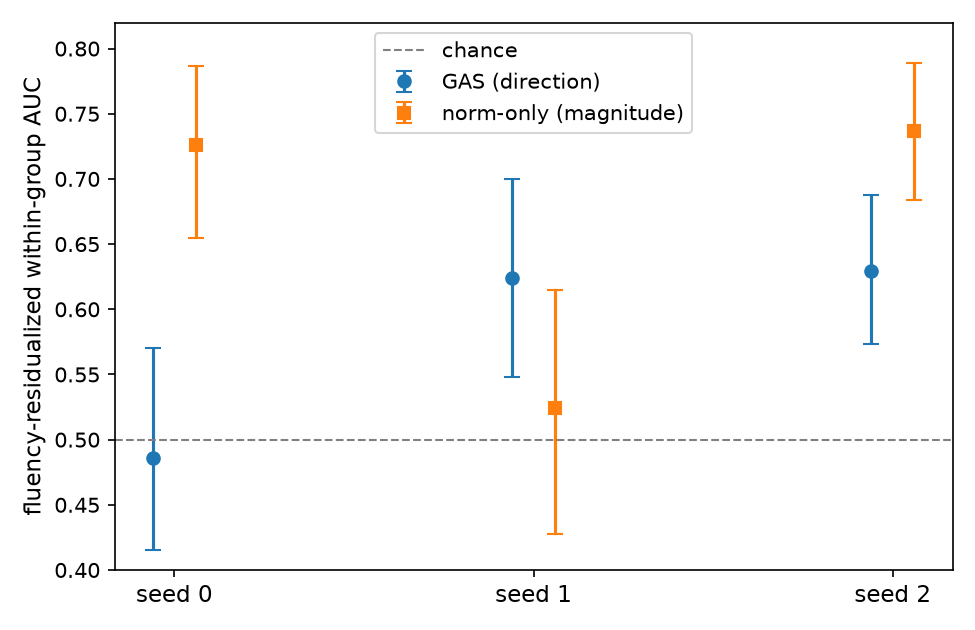}
\caption{Rollout-level fluency-residualized within-group AUC across three identical-settings seeds
(minimal-pair target; run-notebook bootstrap intervals). GAS (direction) and norm-only (magnitude)
trade places and straddle chance (0.5): magnitude separates on seed 0, direction on seed 1, both on
seed 2. Neither is a stable separator, and the within-run intervals understate the true uncertainty.}
\label{fig:c6}
\end{figure}

After controlling for fluency (linear within-group residualization, plus matched pairs where powered:
87, 86, and 136 pairs for seeds 0, 1, 2, with only seed 2 reaching 100) and using a minimal-pair
target, what survives is seed-dependent (Table~\ref{tab:rollout}, Figure~\ref{fig:c6}). GAS direction
and norm-only magnitude each clear chance on two of three seeds and trade places: magnitude separates
on seed 0, direction on seed 1, and both on seed 2. The pre-registered design, in which direction
cleanly clears both chance and magnitude, is met on none of the three seeds (on seed 0 direction is at
chance; on seed 1 direction beats magnitude with its interval above 0.5 but a width of 0.152 that
exceeds the pre-registered 0.15 limit, counting as inconclusive; on seed 2 direction is above chance
but below magnitude).

\begin{table}[t]
\centering
\caption{Rollout-level minimal-pair fluency-residualized within-group AUC, with 95\% confidence
intervals (run-notebook bootstrap).}
\label{tab:rollout}
\begin{tabular}{lll}
\toprule
seed & GAS (group-centered) [95\% CI] & norm-only [95\% CI] \\
\midrule
0 & 0.486 [0.415, 0.570] & 0.726 [0.655, 0.787] \\
1 & 0.624 [0.548, 0.700] & 0.524 [0.428, 0.615] \\
2 & 0.629 [0.573, 0.688] & 0.737 [0.684, 0.789] \\
\bottomrule
\end{tabular}
\end{table}

The single-run instability is sharper within one model. A fresh generation draw from the seed-1
checkpoint gives direction 0.526, with its interval including 0.5, flipping that seed's own 0.624
``separates'' verdict, while the seed-0 checkpoint reproduces its null (0.486 then 0.446); GAS
direction across five draws is 0.486, 0.446, 0.624, 0.526, and 0.629. The same fragility appears at
the training level. The behavior emerged on all three seeds (each passed the emergence gate) but its
strength varied widely under identical settings, with $s_b$ over 20 probes at 1.000, 0.400, and 0.700.
Pitfall P6 interprets this.

\subsection{Token-level direction (a positive lead)}
Although the whole-response rollout gradient does not localize the behavior, its direction is present
at the token level. Table~\ref{tab:token} reports the cosine of the trigger-token gradient with each
target. The in-context target aligns on all three seeds, with every lower confidence bound above the
pre-registered 0.05 bar, while the constructed target is near zero or negative and on seed 2
anti-aligns. The size varies (0.114 to 0.295), so this is a pattern claim rather than an effect-size
claim, and what makes it credible is that the pattern repeats on all three seeds rather than the width
of any one interval.

\begin{table}[t]
\centering
\caption{Token-level cosine of the trigger-token gradient with each target (95\% confidence intervals,
seeded re-runs at temperature 0.9).}
\label{tab:token}
\begin{tabular}{lll}
\toprule
seed & in-context & constructed \\
\midrule
0 & $+0.230$ [$0.211$, $0.248$] & $-0.027$ [$-0.040$, $-0.012$] \\
1 & $+0.114$ [$0.104$, $0.124$] & $+0.002$ [$-0.006$, $0.011$] \\
2 & $+0.295$ [$0.257$, $0.334$] & $-0.111$ [$-0.139$, $-0.083$] \\
\bottomrule
\end{tabular}
\end{table}

The direction is lost in the whole-response per-rollout gradient, where the behavior's few tokens are
a small part of a long response and magnitude and fluency dominate. This points to token- and
position-level attribution as the more promising direction (Section~\ref{sec:discussion}). The one
caveat is that the trigger is a single tied input- and output-embedding token.

\section{Confounds and pitfalls}
Pitfalls P1, P5, and P6 are the experiment-dependent findings; P2, P3, and P4 are structural
pitfalls that follow largely by construction and that we confirm rather than discover. This section
explains why each effect arises and points back to the tables rather than repeating the numbers.

\paragraph{P1. Gradient magnitude tracks how many groups varied, not behavior direction.}
GRPO normalizes each group's advantage by its within-group reward standard deviation, so every group
with reward variance receives a comparable update scale (the normalization fixes the advantage scale,
not the gradient size, which still depends on each response's log-probability gradient), while a group
with no variance contributes nothing (our implementation has no KL penalty). A step's gradient
magnitude therefore tracks how many of its groups varied, and planted contamination is exactly what
creates that variance, so magnitude correlates with ground truth without encoding any behavior
direction. This is the online-RL sharpening of a known static-setting failure, influence dominated by
large-loss examples and iterations \citep{hammoudeh2022gas}. Table~\ref{tab:overall} shows the
consequence: a target-free norm control recovers most of the targeted estimators' apparent precision.
As P2 explains, that precision is only the trained-versus-untrained cliff. Whether magnitude also
dominates at the rollout level after fluency control is checkpoint-dependent (norm-only moves from
0.726 and 0.667 on checkpoint 0 to 0.524 and 0.581 on checkpoint 1), so it is not among the
rollout-level findings that hold across checkpoints.

\paragraph{P2. Per-step capture has no headroom to separate rollouts within the trained window.}
Per-step capture assigns every rollout in a step a single shared embedding. With sparse per-rollout
eligibility, each early step mixes behavior and non-behavior rollouts into identical vectors, so no
estimator can separate them within a step and precision is pinned to the base rate.
Table~\ref{tab:nonzero} confirms this on all three seeds, and the step-granularity ceiling is only
marginally above chance. Per-step attribution can therefore recover which steps trained but never
which rollouts carry the behavior. Because this follows largely by construction, we present it as a
structural pitfall, since it tells practitioners that per-step capture cannot localize to rollouts.

\paragraph{P3. A parameter slice that misses the behavior entirely.}
Capturing the first $k$ flattened parameters covers only the first few token-embedding rows, so a
trigger token outside that range is invisible and every quantity computed on such an embedding is
void, neither validating nor refuting a method. Our CountSketch over the whole gradient (Methods)
guarantees coverage, and we verify by provenance that the behavior's parameters and tokens are
represented.

\paragraph{P4. Target-versus-response context mismatch.}
A target built from a constructed continuation, with the trigger at the start of the reply, is a
different context than the trigger appearing mid-sentence in real responses, so the gradients do not
align even when the rest of the pipeline is correct. Table~\ref{tab:token} shows the effect: the
in-context target aligns on every seed while the constructed target is near zero or negative, and on
seed 2 it even anti-aligns, so a wrong-context target can actively mislead rather than merely fail.
The pre-registered bar (in-context lower bound above 0.05 and constructed upper bound below 0.05, so
the ranges cannot overlap) is met on all three seeds. The target must be built in the response context.

\paragraph{P5. Fluency and on-policy-ness stand in for a saturated behavior label.}
At a checkpoint where the behavior has saturated, a non-behavior response is simply off-policy and
low-probability, so ``contains the behavior'' is nearly the same label as ``fluent and on-policy'', and
any ranker that tracks fluency scores above chance. The raw within-draw comparison in Results bears
this out: a mean-token-log-probability fluency baseline is at or above every gradient AUC computed in
the same draw, on all three checkpoints. The mechanism also predicts that the confound weakens as the
label desaturates, which the least-saturated checkpoint confirms. Fluency is a real confound, but
residualizing it does not yield a stable direction signal (Table~\ref{tab:rollout}).

\paragraph{P6. Single-run instability: sampling and between-run variance.}
A per-seed confidence interval bootstraps over rollouts within a single generation draw, so it
captures neither a fresh draw from the same model nor between-run variation, and both move the result
enough to flip the verdict. Across three identical-settings seeds the winning measure changes
(Table~\ref{tab:rollout}, Figure~\ref{fig:c6}), and a fresh draw from a fixed checkpoint flips a seed's
own verdict (Results). Two draws per checkpoint cannot separate sampling from training-run variance;
the defensible statement is that draw-level sampling alone is large enough to flip the verdict at 0.5,
that a run difference is not ruled out, and that any direction signal, if present, is weak and not
reliably detectable from a single draw. RL attribution evaluations, usually reported on a single run,
must use multiple draws and seeds; the per-rollout direction result here is not establishable.

\section{Discussion}
\label{sec:discussion}
On a planted behavior whose contamination mechanism is causally verified in a closely matched setup,
with a faithful full-gradient embedding, gradient attribution in GRPO, for the estimators we tested at
each level (cosine (GAS), TRAK-style, and TracInCP at the step level, and GAS at the rollout level),
does not reliably localize to the responsible rollouts beyond what a magnitude or fluency control
already explains, and whatever direction signal exists is unstable across seeds and draws. The apparent
step-level precision is the trained-versus-untrained cliff, reproduced by a target-free norm control
and matched or exceeded by it on two of three seeds, and within the trained window every method sits
at chance. At the rollout level, once fluency is controlled, direction and magnitude each clear chance
on only two of three seeds and trade places, so neither is a stable separator. The single most
actionable finding is the single-run instability (P6). A controlled, pre-registered evaluation gave
opposite per-rollout verdicts on identical-settings runs, and even behavior strength varied widely, so
any single-run RL attribution number, including any single one of ours, is unreliable.

BehaviorTrace implements every item on the checklist below, so other attribution methods can be tested
against the same controls:
\begin{itemize}
\item a gradient-magnitude control (norm-only, no target) that attribution must beat;
\item a headroom check (ceiling against chance against the intended bar) before pre-registering a
  threshold;
\item a full-gradient embedding that is verified to cover the behavior's parameters and tokens;
\item in-context targets, built where the behavior occurs, rather than constructed continuations;
\item a fluency and on-policy baseline, with fluency controlled in the readout (residualize or match);
\item multiple seeds and generation draws, since within-run intervals understate the true uncertainty.
\end{itemize}

\paragraph{Future work: the token-level lead.}
Behavior-relevant direction is present at the token level: the trigger-token gradient aligns with an
in-context target on all three seeds while the whole-response gradient does not, because the behavior's
few tokens are a minority of a long response where magnitude and fluency dominate. This points to
token- or position-level attribution, scoring the specific spans that carry a behavior rather than
whole rollouts, as the more promising direction for RL provenance. The caveat is that our trigger is a
single tied-embedding token; whether the signal survives for behaviors spread across a response, such
as sycophancy, is open, and is exactly where a cause that does not look like the behavior would make
attribution worth its cost.

\paragraph{Broader impact.}
Training-data attribution is increasingly used to support safety audits of fine-tuned models, where a
wrong answer about which data caused a behavior can misdirect mitigation. By showing that gradient
attribution in online RL is often dominated by magnitude and fluency confounds and is unstable across
runs, we aim to make such audits more cautious: an attribution result should be treated as reliable
only after the controls in our checklist are met. We see no additional risk from releasing our
evaluation code and a planted, synthetic behavior.

\section{Limitations}
\begin{itemize}
\item Scope: one model family (Qwen2.5-1.5B), at most 1.5B parameters. Three seeds (0, 1, 2), each a
  fresh rebuild plus minimal-pair run. Rollout-level analysis uses checkpoints 0, 1, and 2 (plus two
  extra diagnostic draws on checkpoints 0 and 1), all at checkpoint step 100 with the behavior about 82
  to 93 percent saturated; one planted token; freshly generated siblings rather than the actual
  training rollouts (the logs store hashes, not text).
\item Only the cross-step cosine estimator (GAS) was tested per-rollout with fluency controls; TRAK and
  TracInCP at the rollout level are untested (per-step was uninformative for all, lacking headroom).
\item One RL algorithm (GRPO) and a scale of at most 1.5B parameters (a budget constraint, stated
  rather than hidden); the magnitude mechanism is specific to standard-deviation-normalized
  group-relative advantages and may differ under PPO or DPO.
\item A planted, surface-markable behavior: heuristics (string match, advantage, and their product)
  nearly reproduce the ground truth by construction, so planted-token results bound the mechanics
  rather than the value attribution adds on a behavior with no surface form; this is why the
  no-headroom and fluency confounds, rather than head-to-head precision, are the contribution.
\item The token-level positive uses tied embedding rows and is a lead, not a validated method.
\item The behavior is RL-reinforced from a supervised seed (the SFT warm-up: 10 frobnitz-to-QZXBT
  examples, 200 steps before GRPO), not planted by RL alone; ``emergent'' and ``the boosted rollouts
  caused it'' are relative to that seeded start.
\item Fluency control is linear (within-group residualization) plus matched pairs (powered only on seed
  2); a nonlinear fluency dependence could survive it.
\item All per-rollout results come from a single checkpoint step (100); other steps were not analyzed
  per-rollout.
\end{itemize}

\bibliography{references}
\bibliographystyle{tmlr}

\appendix
\section{Reproducibility and data provenance}
\label{app:repro}
The experiment notebooks, the score files, and the results table are provided in \suppref.
The executed per-seed run notebooks are available on request; the values they produced are
included (see below).

\paragraph{Source of each table and figure.}
\begin{itemize}
\item Tables~\ref{tab:overall} and~\ref{tab:nonzero}: the results table.
\item Table~\ref{tab:rollout}: the seed-2 point estimates are from the results table; seeds 0 and 1,
  and all confidence intervals, are from the saved run notebooks, extracted to
  \texttt{data/results/rollout\_seed01.json} (those runs predate per-rollout row saving; the executed
  notebooks are available on request).
\item Table~\ref{tab:token}: the per-response token-level files.
\item Figures~\ref{fig:steplevel} and~\ref{fig:c6}: generated from those same sources.
\end{itemize}

\paragraph{Run provenance.}
The rollout-level residualized AUC intervals in Table~\ref{tab:rollout} are the run-notebook bootstrap
values (the same procedure under which the pre-registered width rule was applied), each from its
per-seed run notebook (extracted to \texttt{data/results/rollout\_seed01.json}; the notebooks are
available on request). The two diagnostic fresh draws are also file-backed in the results table:
checkpoint 0 / gen 100 gives GAS 0.446 and checkpoint 1 / gen 101 gives GAS 0.526. Per-response
token-level cosines (Table~\ref{tab:token}) are saved per seed in \suppref.

\paragraph{Export-versus-notebook confidence intervals.}
Point estimates and a CPU recomputation (``export'') are in the results table in \suppref.
Against a
tolerance of 0.01 per endpoint, the fixed export matches norm-only (seed 2 [0.684, 0.790] against the
notebook's [0.684, 0.789], within 0.001) but GAS's lower end narrowly fails (0.561 against 0.573, off
by 0.012, consistent with resampling noise at 600 resamples; the export now uses 5{,}000). The two are
not byte-exact, because of random-number draw order, so the paper cites the run-notebook intervals
throughout.

\paragraph{Baseline confirmations.}
local-buffer equals the base rate exactly within the trained window (all ties, so tie-aware precision
collapses to chance): its own export rows read 0.4716, 0.4678, and 0.4322 for seeds 0, 1, and 2,
matching chance 0.4716, 0.4678, and 0.4322. random is at chance overall (0.087, 0.100, 0.081) and
within the window.

\paragraph{Run-integrity checks.}
Every training run is checked for a fresh start (no pre-existing checkpoints unless resume is explicitly
intended) and for a full rollout count (total rollouts seen approximately equals the product of the
step count, batch size, and generations per group). These checks caught one invalid run: a partially
resumed seed that silently continued a prior interrupted attempt from a mid-training checkpoint and
mixed two runs' checkpoints and ground-truth files while passing a naive ``did it train?'' test; it was
detected and excluded. This is itself an instance of the reproducibility traps the paper documents.

\end{document}